\documentclass[letterpaper]{article} 
\usepackage{booktabs}
\usepackage{aaai2026}  
\usepackage{times}  
\usepackage{helvet}  
\usepackage{courier}  
\usepackage[hyphens]{url}  
\usepackage{graphicx} 
\usepackage{natbib}  
\usepackage{caption} 
\usepackage{algorithm}
\usepackage{algorithmic}

\usepackage{newfloat}
\usepackage{listings}
\usepackage{amsmath}
\usepackage{amssymb}
\usepackage{multirow}
\DeclareCaptionStyle{ruled}{labelfont=normalfont,labelsep=colon,strut=off} 
\floatstyle{ruled}
\newfloat{listing}{tb}{lst}{}
\floatname{listing}{Listing}

\title{AAAI Press Formatting Instructions \\for Authors Using \LaTeX{} --- A Guide}
\author{
    Ziyang Song$^{*}$,
    Zelin Zang$^{*}$,
    Zuyao Chen,
    Xusheng Liang,
    Dong Yi, \\
    Jinlin Wu$^{\dagger}$,
    Hongbin Liu,
    Jiebo Luo,
    Zhen Lei,
}
\affiliations{
    Centre for Artificial Intelligence and Robotics (CAIR)\\
    Hong Kong Institute of Science \& Innovation\\
    Chinese Academy of Sciences\\
    \{ziyang.song, zelin.zang, zuyao.chen, xusheng.liang, dong.yi, jinlin.wu, hongbin.liu, jiebo.luo\}@cair-cas.org.hk\\
    $^{*}$Equal contribution, $^{\dagger}$Corresponding author
}

\usepackage{bibentry}

\begin{document}
\title{Anatomy-R1: Enhancing Anatomy Reasoning in Multimodal Large Language Models via Anatomical Similarity Curriculum and Group Diversity Augmentation} 
\maketitle

\begin{abstract} 
Multimodal Large Language Models (MLLMs) have achieved impressive progress in natural image reasoning, yet their potential in medical imaging remains underexplored, especially in clinical anatomical surgical images. Anatomy understanding tasks demand precise understanding and clinically coherent answers, which are difficult to achieve due to the complexity of medical data and the scarcity of high-quality expert annotations. These challenges limit the effectiveness of conventional Supervised Fine-Tuning (SFT) strategies. While recent work has demonstrated that Group Relative Policy Optimization (GRPO) can enhance reasoning in MLLMs without relying on large amounts of data, we find two weaknesses that hinder GRPO's reasoning performance in anatomy recognition: 1) knowledge cannot be effectively shared between different anatomical structures, resulting in uneven information gain and preventing the model from converging, and 2) the model quickly converges to a single reasoning path, suppressing the exploration of diverse strategies. To overcome these challenges, we propose two novel methods. First, we implement a progressive learning strategy called Anatomical Similarity Curriculum Learning by controlling question difficulty via the similarity of answer choices, enabling the model to master complex problems incrementally. Second, we utilize question augmentation referred to as Group Diversity Question Augmentation to expand the model's search space for difficult queries, mitigating the tendency to produce uniform responses. Comprehensive experiments on the SGG-VQA and OmniMedVQA benchmarks show our method achieves a significant improvement across the two benchmarks, demonstrating its effectiveness in enhancing the medical reasoning capabilities of MLLMs. The code can be found in \url{https://github.com/tomato996/Anatomy-R1}
\end{abstract}


\section{Introduction}

Accurate surgical anatomical structure identification is a cornerstone of modern medical practice, underpinning the entire process of disease diagnosis, surgical planning, and treatment formulation, and is crucial for ensuring the accuracy of clinical decisions and patient safety. Doctors rely on a deep understanding of human anatomy to accurately interpret medical images, localize lesions, and perform precise surgical procedures\cite{schütz2024frameworkmultimodalmedicalimage, zang2026cellscout, zang2024evnet}. With the rapid advancement of artificial intelligence, multimodal large language models (MLLMs) have demonstrated immense potential~\cite{li2023llavamedtraininglargelanguageandvision, chen2024huatuogptvisioninjectingmedicalvisual,he2024gscogeneralizableaimedicine}. By analyzing and integrating both medical imaging and textual data, MLLMs promise to automate anatomical structure identification, thereby enhancing clinical efficiency and diagnostic precision\cite{zang2026dmtme}.




However, applying MLLMs to the specific domain of anatomical identification is not without its challenges. Currently, the predominant training methodology is Supervised Fine-Tuning (SFT), which has encountered significant bottlenecks in practice. On one hand, the success of SFT is heavily dependent on large volumes of high-quality annotated data~\cite{liu2024bestpracticeslessonslearned}. In the medical field, precise annotation of anatomical structures is an extremely labor-intensive and costly task, requiring substantial time investment from experienced medical experts, which leads to a scarcity of available high-quality datasets\cite{article}. This inherent data scarcity makes models trained via SFT highly susceptible to overfitting, where the model merely memorizes the features of the training samples but fails to generalize to new, unseen data\cite{zang2023boosting}, resulting in poor performance in real-world clinical applications\cite{zang2025deep}.
\begin{figure}
    \centering
    \includegraphics[width=0.66\linewidth]{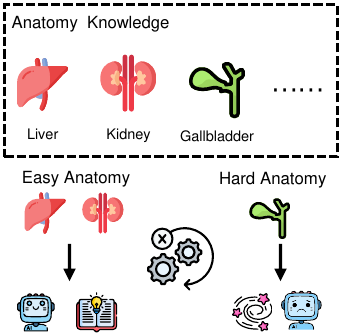}
    \caption{Results of the baseline GRPO model trained without a curriculum. The MLLMs can learn simple anatomies but fail on more challenging ones. Specifically, the knowledge acquired from those simpler structures does not transfer well to the more difficult anatomies.}
    \label{fig:weekness-1}
\end{figure}
To overcome the limitations of SFT in terms of model reasoning capabilities, some researchers have begun to introduce RL into the field of medical artificial intelligence~\cite{zhao2024aquliamed, banerjee2024directpreferenceoptimizationsuppressing}. RL guides the model through an exploratory learning process via a reward mechanism and is believed to enhance the model's complex reasoning abilities, allowing it to move beyond simply mimicking annotated data and thus achieve superior generalization performance compared to SFT. Following this trend, advanced RL algorithms such as Group Relative Policy Optimization (GRPO) have garnered attention due to their efficiency, stemming from the absence of a critic model~\cite{deepseekai2025deepseekr1incentivizingreasoningcapability}. Although GRPO shows promise in certain respects, it faces severe challenges of its own, particularly its frequent exhibition of significant instability during training, such as advantage reward vanish~\cite{park2025deepvideor1videoreinforcementfinetuning, wang2025vlrethinkerincentivizingselfreflectionvisionlanguage} and non-convergence issues\cite{pan2025medvlmr1incentivizingmedicalreasoning}. This makes the model's learning process unpredictable and difficult to reproduce, severely hindering its application in high-reliability medical scenarios.

Therefore, the core challenge for MLLMs in the domain of anatomical learning currently lies in how to ensure the stability and reliability of the model training process. To address this issue, we propose two novel methods aimed at resolving the instability problems encountered by existing reinforcement learning frameworks when training MLLMs for anatomical structure identification, thereby developing more robust and reliable intelligent models to promote their safe application in clinical practice.

First, as shown in Figure \ref{fig:weekness-1}, the knowledge gained from the easy anatomies does not effectively transfer to or enhance the model's comprehension of hard anatomies. This disconnect in knowledge acquisition ultimately prevents the model from achieving stable convergence during the later stages of training. Therefore, inspired by Kimi k1.5~\cite {kimiteam2025kimik15scalingreinforcement}, we introduce a method called \textbf{Anatomical Similarity Curriculum Learning}(ASC-Learning) to control instance difficulty by leveraging the similarity between different options within a single instance. By quantifying these intra-instance similarities, we create a precise, numerical measure of difficulty. For example, the challenge of differentiating the gallbladder from the abdominal wall is substantially greater than that of separating the gallbladder from cystic plate. This quantification allows us to systematically adjust the difficulty of training samples during the RL process, starting with low-similarity (easier) instances and progressively introducing high-similarity (harder) ones as the agent's competence grows. This approach enables fine-grained control over the curriculum, ensuring that each anatomy provides a stable learning signal and preventing a situation where the MLLMs repeatedly fail on more difficult anatomies, thereby ensuring consistent knowledge acquisition.

\begin{figure}
    \centering
    \includegraphics[width=1.0\linewidth]{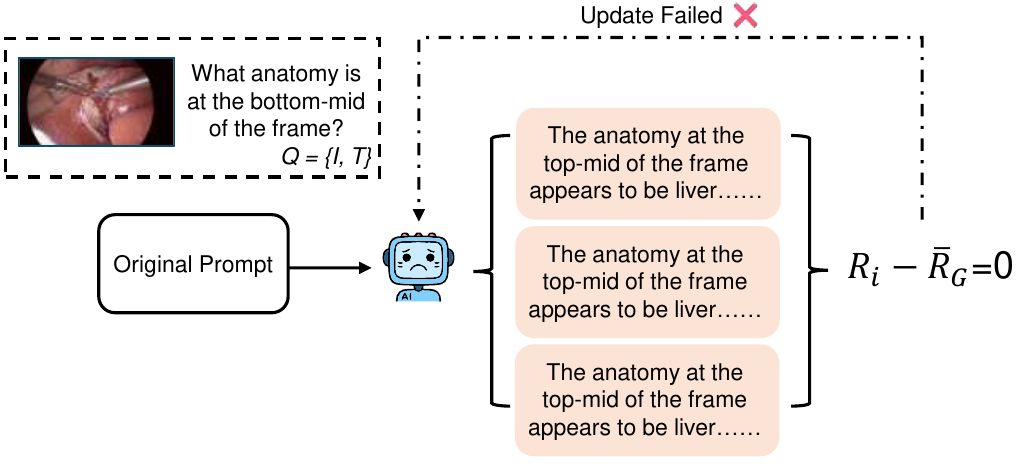}
    \caption{Model responses after several training steps with the direct application of GRPO. The uniformity of the responses indicates that the model's reasoning paths have become restricted. This consequently causes the gradient in this group to become zero, preventing effective updates.}
    \label{fig:weekness2}
\end{figure}

Second, as shown in Figure \ref{fig:weekness2}, the model's responses become uniform, indicating its rapid convergence to a single reasoning path, which greatly hinders the further development of its reasoning capabilities. As training continues, this over-reinforcement of one trajectory systematically suppresses the exploration of alternative reasoning strategies. In this situation, the GRPO algorithm becomes ineffective, as it cannot derive a meaningful learning signal from a group of uniformly incorrect responses, thereby trapping the model on an incorrect convergence trajectory and preventing it from acquiring the diverse reasoning skills needed for complex problems.

To overcome this problem, we introduce a query rewriting strategy referred to as \textbf{Group Diversity Question Augmentation} (GDQA). Instead of relying on a static, canonical query for each anatomical identification, our method transforms each prompt into semantically equivalent, targeted variants. This process effectively broadens the model's reasoning pathways and introduces greater diversity into its inference. By doing so, our method prevents the model from prematurely converging on a single, uniform path of reasoning. This could help to mitigate the vanishing gradient phenomenon. Consequently, the model is empowered to explore a wider solution space, especially when confronted with data from domains where pre-training is scarce, ultimately enhancing its overall reasoning ability.
Our main contributions are as follows:

\begin{enumerate}
    \item We introduce a novel strategy that modulates task difficulty by leveraging intra-instance option similarity. This enables fine-grained control over the training curriculum.

    \item We further introduce a query rewriting diversification strategy. By transforming static queries into semantically equivalent variants, this strategy broadens the model's reasoning pathways and exploratory space.

    \item We validate the effectiveness of our proposed methods through extensive experiments. The results demonstrate that our two strategies significantly improve both the training stability and the final performance of MLLMs for anatomical structure identification.
\end{enumerate}
\section{Related Work}
\subsection{Reasoning in MLLMs}
Previous works have attempted to integrate RL into MLLMs to enhance their reasoning capabilities~\cite{sun2023aligninglargemultimodalmodels, yu2024rlhfvtrustworthymllmsbehavior}. For instance, introduced enhancements to RL training methods, such as modifications to preference optimization (PO), thereby improving the model's reasoning abilities. However, these approaches still depend on external models and require substantial computational resources.

Building on the success of the GRPO algorithm in textual domains, as demonstrated by DEEPSEEK-R1\cite{deepseekai2025deepseekr1incentivizingreasoningcapability}, several studies have extended this training paradigm to multimodal settings. Notably, Visual-RFT~\cite{liu2025visualrftvisualreinforcementfinetuning} pioneered the use of GRPO to strengthen the model's reasoning capabilities, facilitating few-shot learning. Following this, numerous works~\cite{huang2025visionr1incentivizingreasoningcapability, zhou2025r1zerosahamomentvisual, yang2025r1onevisionadvancinggeneralizedmultimodal} have explored GRPO's effectiveness in enhancing the reasoning ability of MLLMs. Furthermore, recent efforts have also recognized the impact of group diversity deficiency on GRPO training. For instance, Decoupled Clip and Dynamic Sampling Policy Optimization(DAPO)~\cite{yu2025dapoopensourcellmreinforcement} incorporates dynamic sampling strategies and other policies, while Group Policy Gradient (GPG)~\cite{chu2025gpgsimplestrongreinforcement} addresses this issue through directly optimizing the original RL objective, and we will compare these approaches in our experiments. Nevertheless, these investigations have primarily concentrated on standard mathematical and coding tasks, neglecting the exploration of reasoning capabilities in other fields. Therefore, this paper focuses on the challenging area of clinical anatomical recognition in medicine.
\subsection{Medical VLMs with RL}
In the medical domain, some research has applied GRPO training to bolster the reasoning abilities of MLLMs. For example, MedVLM-R1~\cite{pan2025medvlmr1incentivizingmedicalreasoning} directly employed GRPO on the OmniMedVQA~\cite{hu2024omnimedvqanewlargescalecomprehensive} dataset for initial exploration. However, they faced convergence difficulties with pathology and OCT images, underscoring the challenges in adapting GRPO to intricate medical imaging tasks. Furthermore, Patho-R1~\cite{zhang2025pathor1multimodalreinforcementlearningbased} activated the model's reasoning through extensive pre-training and supplementary techniques, but the limited availability of medical clinical anatomical data presents a significant obstacle. To overcome these hurdles, particularly the issues of data scarcity and convergence in medical applications, we propose two innovative methods to improve GRPO training for medical MLLMs.
\section{Preliminary}
\subsection{Group Relative Policy Optimization} 
GRPO is a variant of PPO, designed to enhance the performance of LLMs on complex reasoning tasks, such as mathematical and scientific reasoning. Starting with a pretrained MLLM to be optimized, GRPO first uses it to initialize a policy model $\pi_\theta$ and a reference model $\pi_{\text{old}}$. For a given image-text pair $(I, T)$, the reference policy model $\pi_{\theta_{\text{old}}}$ generates a set of responses $\{o_1, o_2, \ldots, o_G\}$. A group-based reward function then computes the corresponding rewards $\{R_1, R_2, \ldots, R_G\}$, which are subsequently used to estimate the advantage $\hat{A}_i$ for each response relative to the group:
\begin{equation}
\label{Advantage}
\hat{A}_i = \frac{R_i - \text{mean}\left( \{R_i\}_{i=1}^G \right)}{\text{std}\left( \{R_i\}_{i=1}^G \right)}.
\end{equation}
The policy model is then optimized by maximizing the following KL objective:
\begin{align}
\mathcal{J}_{\text{GRPO}}(\theta) = 
&\ \mathbb{E}_{q \sim \mathrm{P}(Q), \{o_i\}_{i=1}^{G} \sim \pi_{\theta_{\text{old}}}(O \mid q)}  \notag \\
&\quad \frac{1}{G} \sum_{i=1}^{G} \Bigg[ 
    \min \Bigg( 
        \frac{\pi_{\theta_{\text{new}}}(o_i \mid q)}{\pi_{\theta_{\text{old}}}(o_i \mid q)} A_i, \notag \\
&\qquad \operatorname{clip} \left( 
        \frac{\pi_{\theta_{\text{new}}}(o_i \mid q)}{\pi_{\theta_{\text{old}}}(o_i \mid q)}, 
        1 - \epsilon, 1 + \epsilon 
    \right) A_i 
    \Bigg) \notag \\
&\quad\quad - \beta\, \mathrm{D}_{\mathrm{KL}} \Big( \pi_{\theta_{\text{new}}} \Big\| \pi_{\text{ref}} \Big) 
\Bigg] \label{eq:grpo}
\end{align} 

where $\pi_\theta$ and $\pi_{old}$ are the current and old policy, and $\epsilon$ and $\beta$ are hyper-parameters introduced in PPO.

\section{Method}
\begin{figure}
    \centering
    \includegraphics[width=1.0\linewidth]{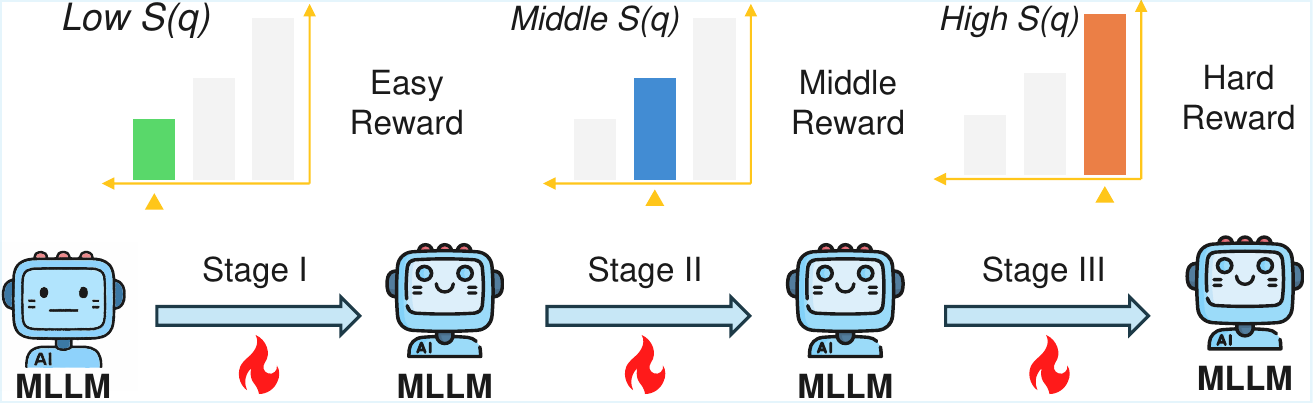}
    \caption{The process of Anatomical Similarity Curriculum Learning. We progressively increase task complexity with aligned reward mechanisms. This enables the model to better acquire knowledge across anatomies of varying difficulty, thereby improving its performance.}
    \label{fig:method1}
\end{figure}
\subsection{Anatomical Similarity Curriculum Learning} 
Previous studies in medical domains, particularly within surgical contexts, have encountered significant challenges in achieving convergence on complex problems, often requiring large-scale pre-training and Chain-of-Thought Supervised Fine-Tuning (CoT-SFT) to stabilize learning. However, such data and resources are scarce in surgical environments, limiting the applicability of these methods. 

To address these limitations, we draw upon progressive learning techniques that have proven effective in code and mathematics domains. These methods involve gradually escalating task complexity to stabilize learning. However, unlike common mathematical tasks where difficulty can be naturally defined (e.g., by the number of steps required to solve a problem), in anatomical recognition tasks, this difficulty is typically hard to define due to the inherent complexity and variability of medical data. Therefore, we propose a novel approach called \textbf{Anatomical Similarity Curriculum Learning} (ASC-Learning) to modulate the difficulty of these tasks by controlling the maximum similarity between alternative options (distractors) and the correct answer in classification tasks. By adjusting this similarity—where lower similarity creates easier tasks with more distinct correct answers and higher similarity increases the challenge by making options more confusable—we can systematically tailor a progressive learning curriculum for anatomical recognition. This method enables stable and efficient convergence without the need for extensive pre-training or abundant data, offering a critical advantage in resource-scarce surgical environments. 

In generating candidate options, we utilize the text encoder component of MedCLIP~\cite{wang2022medclipcontrastivelearningunpaired} to obtain the corresponding embeddings. We then compute the cosine similarity between these embeddings:

\begin{equation} \operatorname{sim}(v_c, v_i) = \frac{v_c \cdot v_i}{|v_c| |v_i|}. \end{equation}

The overall difficulty score S(q) for a question is defined as the maximum similarity value among all distractors, as this represents the most challenging alternative for the model:

\begin{equation} S(q) = \max_{d_i \in D} \operatorname{sim}(E(o_c), E(d_i)). \end{equation}

A low S(q) indicates that all distractors are semantically distinct, making the question easier, while a high S(q) suggests the presence of at least one highly plausible distractor, thereby increasing the difficulty. Our curriculum is structured by partitioning the training data into discrete bins based on these difficulty scores. Training begins with questions from the lowest-difficulty bin and progressively advances to higher-difficulty bins as the model's performance meets predefined thresholds. This graduated approach ensures that the model first learns broad conceptual differences before being required to master the fine-grained distinctions necessary for expert-level anatomical recognition. The progress is shown in Figure \ref{fig:method1}.

\subsection{Group Diversity Question Augmentation} 
Our methodology addresses a critical limitation we identified when applying policy optimization methods such as GRPO to the complex task of surgical anatomy recognition. The core problem stems from a specific form of advantage gradient vanishing that arises from imbalanced learning rates across different anatomical structures.
\begin{figure*}
    \centering
    \includegraphics[width=1.0\linewidth]{AnonymousSubmission/Figure/method2.pdf}
    \caption{The framework of Group Diversity Question Augmentation, where $R_i$ is the MLLM's response to the i-th prompt, and $\overline{R}_G$ is the group-average response. We expand the question space through semantically equivalent transformations and subsequently explore diverse reasoning trajectories arising from different question variants.}
    \label{fig:method-2}
\end{figure*}
Within a single rollout batch, the policy, $\pi_\theta(r_i|s, q)$, generates a group of responses $G(s, q) = \{r_1, \ldots, r_M\}$. Due to the model's initial capabilities and biases, it often learns to recognize common anatomies far more quickly than rarer, more complex structures. This causes the results within the batch to become very uneven:

\begin{itemize}
    \item For \textbf{easy-to-recognize anatomy}, the policy rapidly converges and may correctly identify the structure in all $M$ responses.
    \item For \textbf{difficult-to-recognize anatomy}, the policy consistently fails, incorrectly identifying the structure in all $M$ responses.
\end{itemize}

In both scenarios, the rewards $R_i$ for the responses become homogeneous---either uniformly high (for the easy anatomy) or uniformly low (for the hard anatomy). Consequently, the group-relative advantage \eqref{Advantage} approaches zero for all members of the group concerning that specific anatomical query. This creates a situation where the \textbf{advantage gradient vanishes}, stalling the model's ability to update and improve on these crucial recognitions. The model becomes stuck, having perfected simple tasks but becoming unable to learn complex ones.

To address the learning bottleneck arising from limited question diversity, we propose \textbf{Group Diversity Question Augmentation (GDQA)}, a method inspired by semantically consistent question space expansion\cite{zang2024boosting}. As illustrated in Figure X, given an original multimodal prompt \( Q_0 = \{I, T\} \), where \( I \) is the input image and \( T \) is the textual question, we generate a set of expanded prompts by applying both \textbf{textual} and \textbf{visual semantically consistent transformations}.

As shown in Figure \ref{fig:method-2}, we first employ a textual transformation operator \( \phi(\cdot) \), which rewrites the original prompt into multiple semantically equivalent but syntactically diverse questions. These rewrites retain the original intent and answer, but shift the focus or phrasing (e.g., "Which anatomical structures are located at the lower-center of the image?" or "Can you identify the anatomy present in the central lower portion of the picture?"). This expansion encourages the model to attend to finer-grained or more challenging aspects of the image. In parallel, we introduce a visual transformation operator \( \psi(\cdot) \) that applies controlled modifications to the image (e.g., noise injection) while preserving critical visual cues necessary for reasoning. Each textual variant may be paired with a visually transformed image, resulting in a set of multimodal prompt variants.

For each expanded prompt, the policy \( \pi_\theta(r_i \vert I, T) \) generates a response. By inducing diversity in the queries, our method intentionally increases the variance within the response group \( G(I, T) \), leading to a broader and more informative reward distribution \( R_i \).

\section{Experiment}
\subsection{Implementation Details}
\label{details}
In this work, we adopt Qwen2.5-VL-3B and Qwen2.5-VL-7B~\cite{bai2025qwen25vltechnicalreport} as our base models. We utilized anatomy-related questions from the SGG-VQA dataset for our study. Given that MedVLM-R1~\cite{pan2025medvlmr1incentivizingmedicalreasoning} reported non-convergence on the MI task, we opted to use the MI-specific subset of the OmniMedVQA dataset for our experiments. A detailed introduction to the datasets can be found in the appendix. We employed the MS-SWIFT\footnote{https://github.com/modelscope/ms-swift} framework as the foundation for our reinforcement learning training code, utilizing four A800 GPUs throughout the training process.

To verify the model’s reproducibility, reliability, and robustness on medical questions, we evaluate the model's performance through a suite of metrics: \textbf{Avg@5}, \textbf{Pass@1}, and \textbf{Majority@5}. These metrics are chosen to provide a comprehensive assessment of the model's ability to generate clinically relevant and accurate answers.
Given a medical image \(I\), a question \(Q\), and a set of validated answers, we generate \(k=5\) distinct responses, \(\{ \hat{y}_1, \hat{y}_2, \dots, \hat{y}_5 \}\). The metrics are defined as follows:
\begin{itemize}
    \item \textbf{Avg@5}: Each of the 5 generated responses is scored by a clinical expert on a predefined scale (e.g., from 1 to 5, based on correctness and relevance). The Avg@5 score is the average of these 5 scores, providing an overall measure of response quality.

    \item \textbf{Major@5}: From the 5 generated responses, we identify the most frequently occurring answer through majority voting. This consensus answer is then compared to the ground truth to determine its correctness. This metric assesses the model's consistency and its ability to converge on a correct answer.

    \item \textbf{Pass@k}: This metric measures whether at least one of the top \(k\) generated answers is correct. The formal definition is:
    \begin{equation} 
    \text{Pass@k} = \frac{1}{N} \sum_{i=1}^{N} \mathbb{I}(\exists j \in \{1, \dots, k\} : \hat{y}_{i,j} = y_i),
    \end{equation}
    where \(N\) is the total number of test cases, \(\hat{y}_{i,j}\) is the \(j\)-th generated answer for the \(i\)-th test case, \(y_i\) is the ground truth answer, and \(\mathbb{I}\) is the indicator function.
\end{itemize}
In the context of medical applications, we argue that a metric like \textbf{Pass@5} does not adequately reflect a model's clinical utility. The presence of a correct answer among four incorrect ones is insufficient and potentially hazardous, as a clinician requires the single most reliable answer. A high Pass@5 score could mask a model's inability to prioritize the correct diagnosis or information. Consequently, we exclusively adopt \textbf{Pass@1}, which is equivalent to standard accuracy, to ensure the model's top-ranked prediction is correct and trustworthy. This stricter evaluation aligns with the high-stakes nature of medical decision-making.

\subsection{Main Results}

\begin{table*}[htbp]
\centering
\small
\begin{tabular}{clcccccc}
\toprule
\textbf{Model} & \textbf{Method} & \multicolumn{3}{c}{\textbf{SGG-VQA(Anatomy)}} & \multicolumn{3}{c}{\textbf{OmniMedVQA(MI)}} \\
\cmidrule(lr){3-5} \cmidrule(lr){6-8}
& & \textbf{Avg@5} & \textbf{Pass@1} & \textbf{Major@5} & \textbf{Avg@5} & \textbf{Pass@1} & \textbf{Major@5} \\
\midrule
\multirow{12}{*}{\textbf{Qwen-2.5-VL-3B}} 
&  Zero-Shot             & 26.70 & 25.90 & 26.90 & 60.08 & 59.39 & 68.01\\
&  SFT                   & 34.10 & 35.90 & 32.10 & 77.39 & 77.39 & 77.97 \\
&  CoT-SFT~\cite{wei2023chainofthoughtpromptingelicitsreasoning}               & 26.32 & 24.70 & 25.80 & 68.93 & 70.74 & 71.02\\
\cmidrule(lr){2-8}
&  SFT+GRPO~\cite{chu2025sftmemorizesrlgeneralizes}              & 33.64 & 36.30 & 33.20 & 76.32 & \underline{78.91} & 79.32 \\
&  CoT SFT+GRPO~\cite{li2025relationr1progressivelycognitivechainofthought}          & 31.32 & 34.60 & 34.20 & 67.43 & 68.31 & 68.45\\
\cmidrule(lr){2-8}
& GRPO~\cite{shao2024deepseekmathpushinglimitsmathematical}   & 36.50 & 36.10 & 37.20 & 75.13 & 75.16 & 76.25 \\
&  GPG~\cite{chu2025gpgsimplestrongreinforcement}                   & 40.17 & 41.40 & \underline{40.40} & 76.91 & 78.86 & 78.43\\ 
&  DAPO~\cite{yu2025dapoopensourcellmreinforcement}                  & \underline{40.28} & 42.40 & 39.20 & \underline{77.84} & 78.89 & \underline{80.12}\\
&  GRPO-ASC-Learning     & 40.19 & \underline{42.70} & 40.10 & - & - & -\\
& GRPO-GDQA             & \textbf{43.36} & \textbf{42.80} & \textbf{43.50} & \textbf{79.81} & \textbf{82.38} & \textbf{82.77} \\
\midrule
\multirow{12}{*}{\textbf{Qwen-2.5-VL-7B}} 
&  Zero-Shot             & 29.00 & 29.60 & 29.10 & 60.77 & 62.26 & 69.16 \\
&  SFT                   & 36.46 & 36.10 & 38.40 & 79.16 & 80.27 & 80.08   \\
&  CoT-SFT~\cite{wei2023chainofthoughtpromptingelicitsreasoning}               & 31.25 & 33.40 & 33.70 & 74.13 & 77.78 & 78.31\\
\cmidrule(lr){2-8}
&  SFT+GRPO~\cite{chu2025sftmemorizesrlgeneralizes}              & 41.28 & 42.30 & 42.90 & 77.91 & 79.83 & 80.06\\
&  CoT SFT+GRPO~\cite{li2025relationr1progressivelycognitivechainofthought}          & 34.78 & 33.20 & 34.20 & 76.29 & 76.01 & 77.49\\
\cmidrule(lr){2-8}
&  GRPO~\cite{shao2024deepseekmathpushinglimitsmathematical}             & 38.82 & 39.70 & 38.50 & 76.14 & 77.80 & 80.72\\
&  GPG~\cite{chu2025gpgsimplestrongreinforcement}                   & 41.09 & 40.10 & 41.20 & \underline{80.89} & 80.65 & 81.07\\ 
&  DAPO~\cite{yu2025dapoopensourcellmreinforcement}                  & 40.28 & 39.90 & 40.70 & 80.16 & \underline{81.10} & \underline{82.00} \\
&  GRPO-ASC-Learning     & \underline{42.18} & \underline{43.80} & \underline{44.20} & - & - & -\\
&  GRPO-GDQA             & \textbf{47.32} & \textbf{47.10} & \textbf{47.30} & \textbf{83.60} & \textbf{83.72} & \textbf{83.52} \\
\bottomrule
\end{tabular}
\caption{Performance comparison on the SGG-VQA(Anatomy) and OmniMedVQA(MI) datasets. The implementation details of the methods mentioned in the table can be found in the Appendix. Methods are grouped for clarity. Bold indicates the best performance while underline indicates the second best performance. }
\label{main_results}
\end{table*}

\textbf{Comparison with SFT.} Table \ref{main_results} presents the performance of various methods on the SGG-VQA and OmniMedVQA benchmarks with both 3B and 7B model configurations. A comparative analysis between RL and SFT reveals that RL outperforms SFT across most evaluation results. For instance, in the SGG-VQA (Anatomy) dataset, our RL method achieves an average score of 43.36\% and a Pass@1 of 42.80\% for the 3B model, compared to the SFT's average score of 38.55\% and Pass@1 of 38.00\%. Similarly, in OmniMedVQA, RL improves the Majority@5 score from 67.28\% (SFT) to 71.80\% for the 3B model. For the 7B model, the RL method further demonstrates its superiority. On SGG-VQA, RL achieves an average score of 47.30\%, surpassing SFT's 44.05\%, and significantly enhances the Pass@1 metric from 43.90\% (SFT) to 46.80\%. In OmniMedVQA, RL delivers a Majority@5 score improvement from 72.34\% (SFT) to 76.26\%. \par
These consistent improvements highlight the effectiveness of RL in advancing medical reasoning capabilities, as it leverages iterative feedback to refine understanding and decision-making in complex medical tasks. In summary, compared to direct fine-tuning, the RL-based approach demonstrates substantial performance gains across both benchmarks, underscoring its potential for enhancing medical reasoning and comprehension in visual question answering tasks.\\
\begin{figure}[t]
    \centering
    \includegraphics[width=1.0\linewidth]{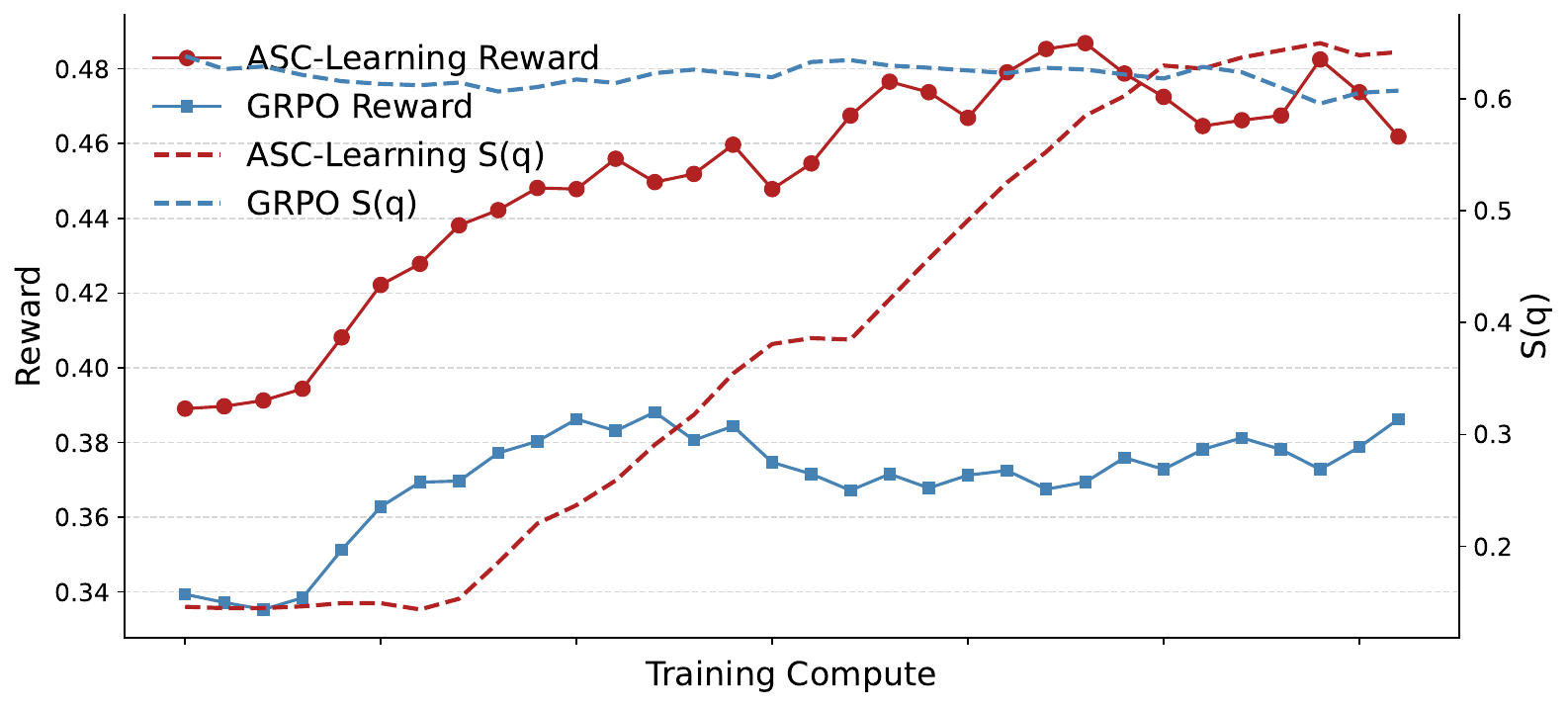}
    \caption{The reward and S(q) curves of the Qwen-2.5-VL-7B during training on the SGG-VQA dataset.}
    \label{fig:analysis method1}
\end{figure}
\textbf{Comparison with other RL-based methods:} Furthermore, when compared to other RL learning strategies, our method demonstrates more significant performance gains. On the SGG-VQA dataset with the Qwen-2.5-VL-7B model, our approach achieves an Avg@5 score of 47.32\%, a substantial 10.86\% improvement over the SFT baseline. In contrast, other competitive RL methods like GPG, while also effective, yield a smaller increase of 4.63\% (41.09\% Avg@5). This trend of superiority is also observed on the OmniMedVQA dataset, where our method outperforms other RL techniques.

We attribute this advantage to the more fundamental manner in which our proposed method addresses the lack of intra-group diversity in generated responses, an issue also targeted by existing approaches like DAPO and GPG. These methods attempt to mitigate this phenomenon through superficial corrections. DAPO, for instance, repeatedly samples a group with a standard deviation of zero until a non-zero standard deviation is achieved. GPG, on the other hand, introduces a scaling factor to amplify the influence of groups within a batch that do not have a standard deviation of zero.

However, our analysis reveals that these methods fail to resolve the underlying problem. We observed that for responses related to difficult anatomical concepts, the model's outputs tend to converge extremely rapidly, often early in the training process. This premature convergence severely diminishes intra-group diversity. In such scenarios, DAPO's strategy of repeated generation becomes ineffective, as the model consistently produces identical responses. Similarly, when faced with a group where the standard deviation is zero, GPG is incapable of extracting a useful learning signal. Although this prevents the gradient estimation from being corrupted by such groups, merely amplifying the gradients of other, non-zero standard deviation groups does not provide the model with the necessary feedback to learn from the more challenging anatomical examples. Consequently, the model fails to receive effective supervisory signals for these difficult-to-learn concepts, hindering its ability to master them.
\begin{figure*}[t]
    \centering 
    \includegraphics[width=1.0\linewidth]{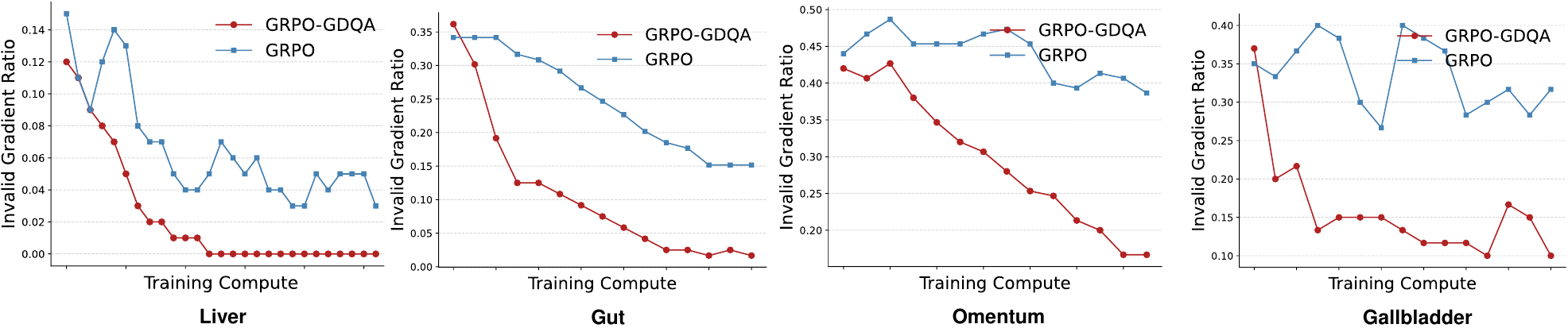}
    \caption{Comparison of invalid gradient ratios between GRPO-GDQA and GRPO across training steps using Qwen-2.5-VL-7B on the SGG-VQA dataset. We define a group as invalid if all responses within it are incorrect. Here, we present the changes in the ratio of ineffective gradient groups from various anatomies across a batch as training progresses.}
    \label{fig:invalid}
\end{figure*}
Conversely, our method addresses this issue by intervening at the input stage. Through the programmatic rewriting of the input prompt, we guide the model to explore broader reasoning paths. This diversification at the input level fosters greater variety in the generated content, thereby preventing the premature convergence to a single response pattern that plagues other approaches. This approach thus offers a more robust and principled resolution to the challenge of maintaining group diversity.\\
\textbf{Comparison between models with different parameter sizes.}
The results also underscore the significant impact of model scale on the efficacy of our proposed reinforcement learning method. As detailed in the table, the performance improvements conferred by our approach over the Zero-Shot baseline are consistently more pronounced on the larger 7B model than on its 3B counterpart across both datasets.
Specifically, on the SGG-VQA (Anatomy) benchmark, our method elevates the Avg@5 score of the Qwen-2.5-VL-7B model by 18.32 percentage points (from 29.00\% to 47.32\%). This surpasses the already substantial 16.66 percentage point gain observed with the 3B model (from 26.70\% to 43.36\%). A similar, even more distinct trend is evident in the OmniMedVQA (MI) results. Here, our method boosts the 7B model’s performance by a remarkable 22.83 percentage points (from 60.77\% to 83.60\%), compared to a 19.73 percentage point increase for the 3B model (from 60.08\% to 79.81\%). This pattern strongly indicates that the intrinsic capabilities of the foundational model are a key factor in the success of the RL-based fine-tuning; a more powerful base model appears better equipped to leverage the optimization provided by the RL method, leading to more significant gains.




\section{Analysis}

\subsection{The Effectiveness of ASC-Learning} 
To further demonstrate that our ASC-Learning effectively addresses the issue of imbalanced information gain from different anatomical categories during training, as shown in Figure \ref{fig:analysis method1}, it can be seen that ASC-Learning significantly outperforms the GRPO strategy: the model's Reward steadily increases throughout the training process and achieves higher values in the later stages. In contrast, GRPO's Reward stops improving after an initial rise in the early training phase, indicating that while the model gains some improvement from easy anatomy, it is unable to effectively acquire knowledge from the reasoning paths of easy anatomy when faced with hard anatomy. Overall, ASC-Learning effectively addresses the issue of limited reasoning ability in MLLMs caused by imbalanced pre-trained knowledge when reasoning.

\subsection{The Effectiveness of GDQA} 


In Figure \ref{fig:invalid}, we present the trends of invalid gradient proportions for four different anatomies under both GRPO and GRPO-GDQA. For simpler anatomies, such as the Liver, we observe that the invalid gradient proportions for both methods eventually converge; however, GRPO-GDQA achieves faster convergence compared to GRPO. This suggests that broadening the exploration paths via GDQA facilitates better convergence in the case of simple anatomies. For other anatomies, GRPO-GDQA consistently maintains a notably lower proportion of invalid gradients throughout training compared to GRPO. Furthermore, as training progresses, the proportion of invalid gradients in GDQA continues to decrease and eventually stabilizes at a relatively low level, whereas the proportion in GRPO remains higher. These observations indicate that GDQA, by enhancing the diversity among responses within groups, effectively broadens the model’s exploration paths, thereby reducing invalid gradients caused by entirely incorrect groups and mitigating the issue of gradient vanishing. Overall, our findings suggest that GDQA can sustain more informative gradient signals during training, offering the model richer optimization dynamics.

\subsection{The Ablation Study of GDQA} 
\begin{table}[htbp]
  \centering
  \begin{tabular}{l|ccc}
    \toprule
    \textbf{Method} & \textbf{Avg@5} & \textbf{Pass@1} & \textbf{Major@5} \\
    \midrule
    GRPO-GDQA & 47.32\% & 47.10\% & 47.30\% \\
    \midrule
    $-Text$ & 46.21\%  & 46.10\%  & 45.40\%  \\
    \midrule
    $-Image$ &  45.34\%  & 46.30\%  & 46.00\%\\
    \bottomrule
  \end{tabular}
  \caption{The ablation study of GDQA on SGG-VQA, $-Text$ denotes the model without question rewriting, and $-Image$ indicates the model without image augmentation.}
  \label{ablation_moe}
\end{table} 
To validate the effectiveness of our proposed GDQA, we conduct a comprehensive ablation study to analyze the contributions of its key components. Our approach enhances performance by increasing response diversity within the GRPO group through text-based augmentation and image augmentation. As shown in Table \ref{ablation_moe}, removing either the text or image augmentation leads to a clear decline across all metrics. Notably, the full GDQA achieves the highest overall results, while excluding question rewriting or image augmentation results in consistent performance drops. These findings confirm that both strategies are essential and complementary for maximizing the effectiveness of GDQA.
\section{Conclusion}
In this paper, we present Anatomical Similarity Curriculum Learning and Group Diversity Question Augmentation, which together effectively address the critical challenges in applying GRPO to medical MLLMs. By balancing the learning gains between complex and simple medical problems, our approach mitigates the problems of non-convergence and suboptimal performance that arise from the scarcity of high-quality training data in the medical domain. We believe this work will further propel the advancement and development of MLLMs in the medical field, paving the way for more robust and reliable clinical applications.

\section*{Acknowledgments}
This work was supported in part by the National Natural Science Foundation of China (Grant No.\#62306313) and the InnoHK Program by the Hong Kong SAR Government.

\bibliography{aaai2026}
\end{document}